\documentclass[letterpaper, 10 pt, conference]{ieeeconf}  

\IEEEoverridecommandlockouts                              
\usepackage[utf8]{inputenc}
\usepackage[T1]{fontenc}

\usepackage{amsmath} 
\usepackage{amssymb}  
\usepackage{siunitx}
\usepackage{tikz}
\usepackage{hyperref}
\usetikzlibrary{shapes, arrows.meta, positioning}
\usepackage[dvipsnames]{xcolor}

\usepackage{makecell}
\usepackage{todonotes}
\presetkeys{todonotes}{inline,backgroundcolor=yellow}{}

\renewcommand{\matrix}[1]{\mathbf{#1}}

\title{\LARGE \bf
Blinking Beyond EAR: A Stable Eyelid Angle Metric for Driver Drowsiness Detection and Data Augmentatio
}

\author{Mathis Wolter$^{1}$, Julie Stephany Berrio Perez$^{2}$, Mao Shan$^{2}$
\thanks{This research was partially funded by the Research Internships in Science and Engineering (RISE) Worldwide grant from the German Academic Exchange Service (DAAD).}
\thanks{The Australian Research Council provided financial support for this project through the Australian Research Council Industrial Transformation Training Centre for Automated Vehicles in Rural and Remote Regions (IC230100001).)}
\thanks{$^{1}$M. Wolter is with the Institue of Smart Engineering and Machine Elements at the Hamburg University of Technology, Hamburg, Germany
        {\tt\small mathis.wolter at tuhh.de}}%
\thanks{$^{2}$J. S. Berrio Perez and M. Shan are with the Intelligent Transport Systems group at the Australian Center for Robotics, The University of Sydney,
        Sydney, NSW 2006, Australia
        {\tt\small stephany.berrioperez at sydney.edu.au, mao.shan at sydney.edu.au}}%
}

\begin{document}

\maketitle
\thispagestyle{empty}
\pagestyle{empty}

\begin{abstract}

Detecting driver drowsiness reliably is crucial for enhancing road safety and supporting advanced driver assistance systems (ADAS). We introduce the Eyelid Angle (ELA), a novel, reproducible metric of eye openness derived from 3D facial landmarks. Unlike conventional binary eye state estimators or 2D measures, such as the Eye Aspect Ratio (EAR), the ELA provides a stable geometric description of eyelid motion that is robust to variations in camera angle. Using the ELA, we design a blink detection framework that extracts temporal characteristics, including the closing, closed, and reopening durations, which are shown to correlate with drowsiness levels. To address the scarcity and risk of collecting natural drowsiness data, we further leverage ELA signals to animate rigged avatars in Blender 3D, enabling the creation of realistic synthetic datasets with controllable noise, camera viewpoints, and blink dynamics. Experimental results in public driver monitoring datasets demonstrate that the ELA offers lower variance under viewpoint changes compared to EAR and achieves accurate blink detection. At the same time, synthetic augmentation expands the diversity of training data for drowsiness recognition. Our findings highlight the ELA as both a reliable biometric measure and a powerful tool for generating scalable datasets in driver state monitoring.
\texttt{URL: The link with the code will be made publicly available upon acceptance.}
\end{abstract}

\section{INTRODUCTION}


 Driver alertness is a central challenge in road safety, particularly with the increasing adoption of Advanced Driver Assistance Systems (ADAS). Although ADAS reduce the burden of manual control, they can also reduce driver involvement in driving tasks, increasing the risk of drowsiness and inattention \cite{bodaghi_ul-dd_2025}. Drowsy drivers typically require more time to respond to emergency takeover requests and often perform takeovers with lower quality, increasing the likelihood of accidents \cite{caffier_experimental_2003, akerstedt_subjective_1990}. Indeed, the National Highway Traffic Safety Administration (NHTSA) \cite{goel_neurocognitive_2009} estimates that drowsy driving is responsible for thousands of crashes annually in the US alone, highlighting the urgent need for effective monitoring technologies.

A wide range of approaches have been explored to detect driver drowsiness. Physiological signals such as EEG, EOG, or heart rate variability can capture reliable indicators of fatigue \cite{bodaghi_ul-dd_2025, jap_using_2009}, but their invasive nature makes them impractical for deployment in consumer vehicles.
Consequently, non-invasive methods have become the dominant research focus, with many leveraging computer vision (CV) to analyze facial and ocular behavior \cite{baccour_camera-based_2019, ghoddoosian_realistic_2019}. Among these, blink dynamics and eyelid behavior are particularly promising indicators of drowsiness. For example, Caffier \textit{et al.} \cite{caffier_experimental_2003} demonstrated significant correlations between blink duration, reopening time, and subjective sleepiness levels, while others have shown that prolonged eyelid closure or frequent microsleeps are strong predictors of drowsy states \cite{wierwille_evaluation_1994, wierwille_research_1996}.

\begin{figure}
    \centering
    \includegraphics[width=0.99\linewidth]{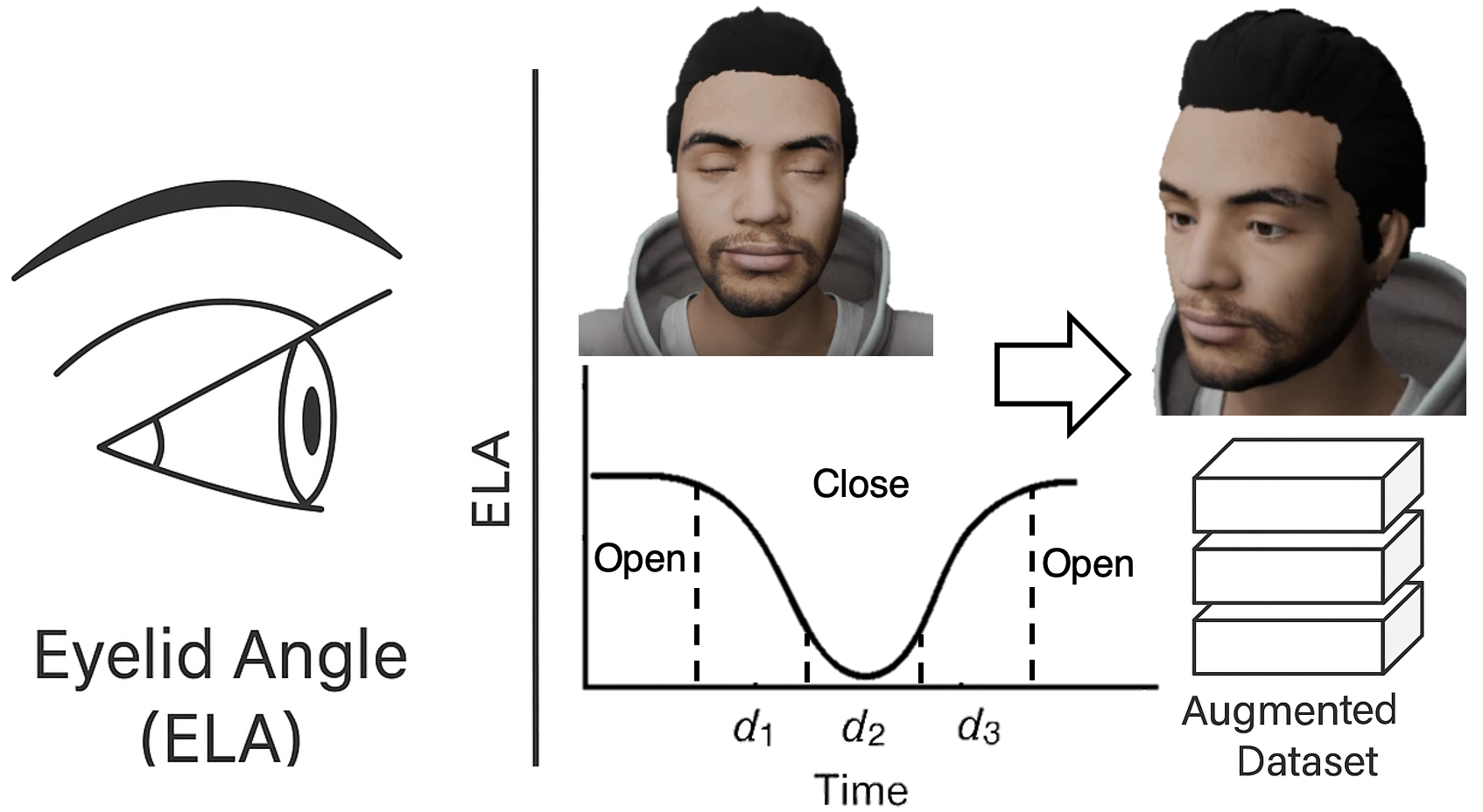}
    \caption{Overview of the proposed Eyelid Angle (ELA) framework: defining eyelid geometry, extracting blink dynamics, and generating synthetic datasets for drowsiness detection.}
    \label{fig:teaser}
\end{figure}


Another critical challenge lies in the scarcity of high-quality, annotated drowsiness datasets. Collecting naturalistic data from drowsy drivers is not only difficult and costly, but also dangerous and ethically constrained \cite{liu_synblink_2023}. Publicly available datasets such as UTA-RLDD \cite{ghoddoosian_realistic_2019}, NTHU-DDD \cite{weng_driver_2017}, and the Driver Monitoring Dataset (DMD) \cite{ortega_dmd_2020} have advanced the field, but their variability in recording setups, frame rates, and annotation quality introduces inconsistencies. 

In this work, we introduce the Eyelid Angle (ELA), a novel, reproducible metric that captures the relative angle between the upper and lower eyelids. Unlike Eye Aspect Ratio (EAR), the ELA is derived from 3D landmark geometry, making it more robust to viewpoint variations. We show how the ELA can be used to accurately detect blinks and extract temporal features (e.g., closing, closed, and reopening durations) that are strongly correlated with drowsiness. Furthermore, we leverage the ELA to generate synthetic data by animating a 3D avatar in Blender, enabling realistic replication and augmentation of blink dynamics. This approach not only provides a stable metric for real-world driver monitoring but also offers a scalable pipeline for creating large, diverse datasets.

The key contributions of this paper are as follows:

\begin{enumerate}
    \item We introduce the ELA as a robust measure of eye openness, invariant to camera viewpoint and suitable for blink analysis.
    \item We demonstrate how the ELA supports the extraction of temporal blink features that correlate with drowsiness and can be used in classification tasks.
    \item We present a pipeline for generating realistic synthetic blink datasets using Blender, driven directly by ELA signals, to augment existing datasets and improve generalization.
    \item We evaluate the proposed method on public driver monitoring datasets and benchmark its robustness against EAR under varying head poses.

\end{enumerate}

By bridging the gap between robust feature extraction and scalable data generation, our approach advances the methodological and practical aspects of driver state monitoring.

\section{BACKGROUND} 



Drowsiness is a complex physiological and cognitive state that affects vigilance and reaction time. It is often defined subjectively using the Karolinska Sleepiness Scale (KSS) \cite{akerstedt_subjective_1990}, although the objective thresholds vary widely between individuals. This high degree of interpersonal variability complicates the design of universal detection systems, which motivates the need for robust and generalizable metrics.


There are two broad categories of drowsiness detection: biometric and CV-based approaches. Biometric methods leverage physiological signals such as EEG, EOG, heart rate variability, or respiration, which can provide precise indicators of fatigue \cite{jap_using_2009, bodaghi_ul-dd_2025}. However, their invasive nature, the reliance on specialized hardware, and low user acceptance limit their feasibility for deployment\cite{lal_driver_2002}
In contrast, CV-based methods are non-invasive and rely on visual cues such as gaze, yawning, head pose \cite{10588436}, or eye behavior \cite {ghoddoosian_realistic_2019, weng_driver_2017, may_driver_2009} 
. Among these, ocular activity, particularly blink dynamics, has been shown to be strongly correlated with drowsiness. Caffier \textit{et al.} \cite{caffier_experimental_2003} reported that blink duration, reopening time, and the proportion of long blinks differ significantly between alert and drowsy states, highlighting the importance of eyelid motion as a behavioral biomarker.


Eye tracking in drowsiness research has been pursued through multiple modalities. Near-infrared imaging is widely used because the pupil and iris exhibit strong contrast in this spectral range, enabling robust detection under varying lighting conditions. For example, Baccour \textit{et al.} \cite{baccour_camera-based_2019} introduced the EyeClosure metric, a normalized measure of iris coverage in infrared images, to quantify eyelid closure.
With the advent of deep learning, RGB-based facial landmark detection has emerged as an attractive alternative. Soukupova and Cech \cite{soukupova_real-time_2016} introduced the EAR, derived from eyelid landmarks, and demonstrated real-time blink detection using a Support Vector Machine classifier. Although simple and computationally efficient, the EAR is inherently viewpoint dependent, as it relies on 2D landmark distances that distort under head rotation \cite{kazemi_one_2014}.
Recent advances in landmark detection, such as MediaPipe Face Mesh \cite{yan_mediapipe_2022} and dlib \cite{kazemi_one_2014}, offer high-density landmarks with improved robustness. Compared to dlib, which provides sparse 2D landmarks, MediaPipe estimates dense 3D facial geometry, enabling more sophisticated metrics of eyelid motion. These advances motivate the search for geometric features that are both reproducible and viewpoint-invariant.

The accuracy of blink-based drowsiness metrics is heavily dependent on video resolution and frame rate. For example, Baccour \textit{et al.} \cite{baccour_camera-based_2019} used recordings at 50 fps and applied a Savitzky–Golay filter \cite{savitzky_smoothing_1964} to smooth the eyelid closure signal before clustering the extrema with k means. Caffier \textit{et al.} \cite{caffier_experimental_2003} instead fitted regression lines to blink flanks, allowing a precise estimation of closing and reopening times. Both studies highlight that higher frame rates significantly reduce temporal error in blink feature extraction.
Based on \cite{baccour_camera-based_2019}, Dreißig \textit{et al.} \cite{dreisig_driver_2020} used a k-Nearest-Neighbors classifier trained on blink characteristics such as reopening duration, amplitude-velocity ratio and blink duration variance. Their results confirmed that blink dynamics can robustly separate alert from drowsy states.


The development of reliable algorithms has been accelerated by the availability of annotated datasets. The UTA-RLDD dataset \cite{ghoddoosian_realistic_2019} provides more than 30 hours of video classified into alert, low-vigilant, and drowsy classes, recorded across different camera types and frame rates (12–30 fps). Similarly, the NTHU-DDD \cite{weng_driver_2017} and DMD \cite{ortega_dmd_2020} datasets include labeled blink and drowsiness events, although simulated drowsiness behaviors raise concerns about ecological validity \cite{ghoddoosian_realistic_2019}.
To address data scarcity, several synthetic datasets have been proposed in the domain of eye and blink detection but not applied to drowsiness detection tasks. SynBlink \cite{liu_synblink_2023} uses Blender 3D to generate artificial blink sequences through interpolations of open and closed frames. UnityEyes \cite{wood_learning_2016, smith_unityeyes_2025}, and its extensions generate large-scale still image datasets for gaze and eye tracking applications. Borges \textit{et al.} \cite{borges_automated_2020} further demonstrated the utility of synthetic in-car datasets for pose estimation. Although valuable, most synthetic datasets lack the fine-grained temporal dynamics of natural blinks, limiting their use for drowsiness analysis.
One of the challenges in the detection of drowsiness is interpersonal variability in blink behavior. Factors such as age, fatigue resistance, and habitual blinking frequency can strongly influence baseline eyelid metrics \cite{nystrom_what_2024}. This variability needs features that are not only stable between individuals but also reproducible across different recording conditions.

The state of the art has established blink behavior as a key indicator of drowsiness, with both biometric and vision-based methods contributing important insights. However, existing visual metrics such as EAR are limited by sensitivity to head pose and camera viewpoint, while synthetic datasets often fail to capture the complexity of natural blink dynamics. These limitations motivate the development of new reproducible and geometrically grounded features, such as the ELA, that can robustly characterize eyelid motion and support both real-world monitoring and synthetic data generation.
\section{METHODOLOGY}



In this section, we introduce the ELA, a novel geometric descriptor that quantifies the relative orientation between the upper and lower eyelids.

To extract facial landmarks, we employ Google’s MediaPipe Face Landmark detector, which integrates the MediaPipe Face Mesh V2 \cite{yan_mediapipe_2022} and the BlazeFace face detector \cite{bazarevsky_blazeface_2019}. Each landmark is estimated in 3D, with the $x$ and $y$ coordinates normalized to the image width and height, respectively. The $z$-axis is defined relative to the centroid of all landmarks and scaled by the facial width.
For consistent coordinate system, we normalize the $y$-values by the image aspect ratio and heuristically rescale the $z$-coordinates as follows:
\begin{align}
    z = 1.7\cdot z_{\mathrm{raw}} \cdot \matrix T_{2,2} \label{eq:zheuristic}    
\end{align}
where $z_\mathrm{raw}$ denotes the raw landmark depth, and $\matrix T$  is the transformation matrix derived from the MediaPipe Face Landmark inference process.
Because a single RGB image lacks inherent depth information, these transformations serve to enforce geometric consistency across landmarks rather than recover true 3D structure.

Most publicly available drowsiness datasets comprise single-view RGB videos, which precludes the use of multi-camera triangulation or other stereo-based methods for 3D reconstruction. Although the DMD dataset \cite{ortega_dmd_2020} includes depth information, its spatial resolution is insufficient for accurately capturing fine eyelid structures.
Compared to alternative landmark detection approaches, such as the dlib facial landmark detector based on \cite{kazemi_one_2014}, MediaPipe offers a denser landmark distribution around the eyelids. Moreover, while dlib estimates only 2D landmark coordinates in image space, MediaPipe provides 3D landmark inference, enabling a richer geometric representation of eyelid motion and orientation.

\subsubsection{Eye features}
The MediaPipe Face Mesh V2 model defines each eyelid using seven landmarks, providing sufficient spatial resolution to characterize eyelid geometry. Using the inferred 3D facial landmarks, we isolate and extract the subset corresponding to the upper and lower eyelids of each eye, which serve as inputs to our proposed algorithm.
To assess the robustness of face detection across a sequence, we introduce the Detection Ratio (DR), defined as the ratio between the number of frames in which a face is successfully detected and the total number of frames in the sequence. This metric provides a quantitative measure of detection reliability under varying conditions such as occlusions, head pose changes, and illumination variations.


\subsubsection{Calculation of the ELA}


To quantify the orientation of each eyelid, we fit a plane to the corresponding 3D landmarks, enabling computation of the relative angle between the upper and lower eyelids. Given a set of $n$ landmarks in three-dimensional space, the objective is to determine the plane that best approximates these points in a least-squares sense.
The landmark coordinates for an eyelid are represented as:

\[
    \matrix L = \begin{pmatrix}
        x_1 &     & x_n \\
        y_1 & ... & y_n\\
        z_1 &     & z_n
    \end{pmatrix}
\]
where each column corresponds to a landmark expressed in normalized coordinates. The centroid of the landmarks is subtracted to obtain a zero-mean matrix 
 \[
    \matrix A = \matrix L - \frac{1}{n}\,\matrix L\,\matrix J_n
\]

where $J_n$ is an $n\times n$ matrix of ones.
A Singular Value Decomposition (SVD)
\[
    \matrix A = \matrix U\,\matrix \Sigma\,\matrix V^T,
\] is applied to find the normal vector of the best fitting plane as 
\[
    \vec n = \matrix U_{*,n}.
\]


Since the direction of $\vec n$ is ambiguous up to a sign, its orientation must be standardized. The eyelid landmarks are ordered from the inner corner (near the nose) to the outer corner of the eye. A coarse directional estimate is computed using the cross product of consecutive landmark vectors: \[
    s = \vec n \cdot (\matrix A_{*,i+1} \times \matrix A_{*,i}).
\] 

If $s < 0$, the normal vector is inverted ($\vec{n} \leftarrow -\vec{n}$) to ensure consistent directionality across eyelids.

\begin{figure}[tb]
    \centering
    \vspace{2mm}
    \includegraphics[trim={0 5cm 0 3.5cm},clip, width=0.95\linewidth]{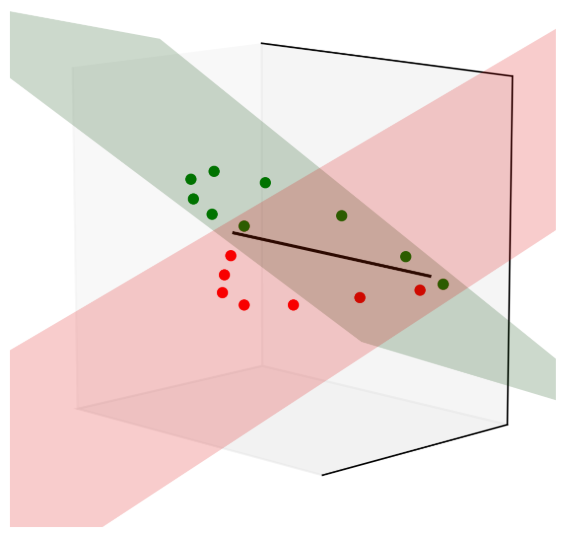}
    \caption{3D landmarks of the upper \textcolor{ForestGreen}{(green)} and lower \textcolor{red}{(red)} eyelids with fitted planes. The relative orientation of these planes defines the ELA, providing a robust geometric representation of eyelid motion from noisy landmark data.}
    \label{fig:surfaces}
\end{figure}

With the normal vectors of the upper and lower eyelids, denoted as $\vec{n}_u$ and $\vec{n}_l$, respectively, the angle between the corresponding planes is computed as
\begin{equation}
    \mathrm{ELA}_{\mathrm{raw}} = \arccos(\vec{n}_{l} \cdot \vec{n}_u),
\end{equation}
where $\mathrm{ELA}_{\mathrm{raw}}$ represents the raw \emph{Eyelid Angle} (ELA) for a single eye. This metric quantifies the relative angular displacement between the upper and lower eyelids. The computation is performed independently for the left and right eyes, resulting in $\mathrm{ELA}_{\mathrm{left}}$ and $\mathrm{ELA}_{\mathrm{right}}$.

To obtain a unified measure that accounts for head pose, we combine the ELA values of both eyes using a visibility-based weighting scheme. Specifically, we apply a sigmoid weighting function $\sigma\left(\cdot\right)$ that depends on the yaw rotation of the face mesh around the vertical $y$-axis, denoted by $\beta$:
\begin{equation}
    \mathrm{ELA}_{\mathrm{combined}} = 
    \sigma\left(-4\beta\right)\,\mathrm{ELA}_{\mathrm{left}} +
    \sigma\left(4\beta\right)\,\mathrm{ELA}_{\mathrm{right}},
\end{equation}
where the factor $4$ is a heuristic scaling term controlling the influence of each side. The rotation parameter $\beta$ is extracted from the face rotation matrix provided by MediaPipe inference. 
Unlike \cite{soukupova_real-time_2016}, which averages the EAR in both eyes, our approach explicitly incorporates visibility weighting to maintain robustness under large head rotations.



\subsubsection{Post processing}\label{sec:postprocessing}
For temporal analysis, the ELA signal is smoothed using a one-dimensional Gaussian filter with a standard deviation proportional to the video frame rate:
\begin{equation}
    \sigma = \frac{\mathrm{FPS}}{30}.
\end{equation}
The filtered signal is then obtained as
\begin{equation}
    \mathrm{ELA}_{\mathrm{filtered}} = \mathrm{Gauss}_\sigma \big(\mathrm{ELA}_{\mathrm{raw}}\big).
\end{equation}
This proportional scaling ensures that videos with higher frame rates exhibit stronger smoothing, effectively reducing jitter, while lower frame rate sequences are not over-smoothed, thereby preserving physiologically relevant eyelid motion.

Although the Savitzky–Golay (SG) filter \cite{savitzky_smoothing_1964} is frequently used for similar purposes \cite{baccour_camera-based_2019, nystrom_what_2024}, we observed that it introduced undesirable artifacts, particularly at low frame rates. In these cases, the SG filter produced pronounced positive spikes immediately preceding the closing phase of a blink. In contrast, the proposed 1D Gaussian filter maintains the natural temporal profile of the signal and better preserves the reopening phase dynamics.

\subsubsection{Detecting blinks}

To detect individual blinks, we employ a signal analysis approach inspired by Baccour \textit{et al.}~\cite{baccour_camera-based_2019}, in which the positive and negative peaks of the derivative of the smoothed ELA signal are clustered using a $k$-means classifier. 
Since blinks exhibit a characteristic temporal pattern, the descending and ascending edges of the signal can be paired to delineate individual blink events~\cite{baccour_camera-based_2019}.

The temporal derivative is computed using a central difference scheme. 
A \emph{blink window} is defined from the nearest local maximum preceding the maximum descent to the first local maximum following the maximum ascent, as illustrated in Fig.~\ref{fig:blinkfeatures}. 
Detected blinks are rejected if the signal exhibits an intermediate between the falling and rising edges that exceeds either the start or end amplitude of the window, as such cases typically correspond to erroneously merged blinks where reopening and reclosing occur too slowly to be classified separately.

Due to the unsupervised nature of the $k$-means clustering, this method is not directly applicable for real-time detection, as it requires a sufficient number of data points for stable classification. 
In our implementation, blinks are analyzed every $\qty{60}{\second}$ using a sliding window containing the most recent $\qty{90}{\second}$ of ELA data.

\subsubsection{Blink features}\label{sec:blinkmetrics}

\begin{figure}[t!]
    \centering
    \vspace{2mm}
    \includegraphics[width=0.98\linewidth]{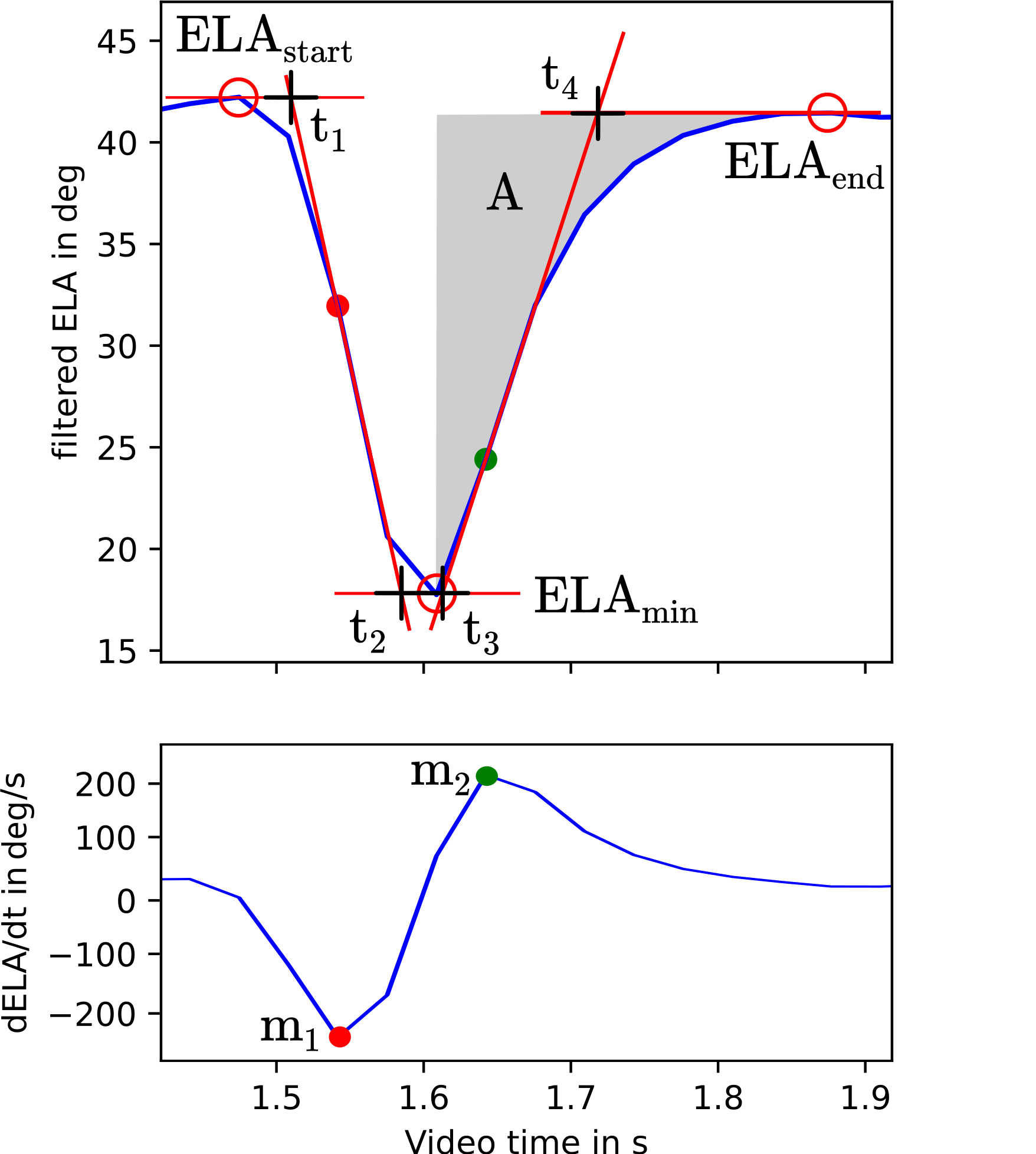}
    \caption{A \emph{blink} is defined as the interval between the last local maximum $\mathrm{ELA}_{\mathrm{start}}$ preceding the minimum derivative $m_1$ and the first local maximum $\mathrm{ELA}_{\mathrm{end}}$ following the maximum derivative $m_2$. 
    The extrema of the derivative, $m_1$ and $m_2$, are classified using a $k$-means algorithm, as described in~\cite{baccour_camera-based_2019}. 
    The intersections of the tangents (shown as red lines) define the temporal boundaries used to compute the \emph{closing}, \emph{closed}, and \emph{reopening} durations of the blink. 
    All features extracted from this process are summarized in Table~\ref{tab:blinkfeatures}. 
    The example shown in Fig.~\ref{fig:blinkfeatures} was recorded using a standard webcam operating at $\qty{30}{\hertz}$.
}\label{fig:blinkfeatures}
\end{figure}

Caffier \textit{et al.}~\cite{caffier_experimental_2003} define several temporal metrics of a blink, namely the \emph{closing}, \emph{closed}, and \emph{reopening} durations. 
In their approach, the closing and reopening phases are determined by applying linear regression (LR) to the respective signal flanks, identifying the intersections of the fitted lines with the baseline and with each other to mark the onset and offset of each phase. 
However, in our case, each flank consists of only a limited number of data points, which leads to numerical instability in the LR fitting process and renders this method unreliable.

To overcome this limitation, we estimate these temporal intervals using the derivative of the filtered ELA signal. 
Four characteristic intersection points are computed to define the blink phases. 
Specifically, the \emph{closing duration} is defined between (i) the intersection of the nearest local maximum before the blink with the tangent fitted through the point of maximum descent, and (ii) the intersection of that tangent with a constant line at the signal minimum. 
The \emph{opening duration} is determined analogously using the tangent through the point of maximum ascent. 
The \emph{closed duration} corresponds to the time interval between the intersections of the descending and ascending tangents with the minimum-level line. 
This procedure is illustrated in Fig.~\ref{fig:blinkfeatures}, and the full set of computed temporal features is summarized in Table~\ref{tab:blinkfeatures}.

\begin{table}[tb]
    \centering
    \caption{Features calculated for each detected blink. Indices and intersections are shown in Fig. \ref{fig:blinkfeatures}}
    \label{tab:blinkfeatures}
    \begin{tabular}{|c || l|} \hline
        \bf{Name} & \bf{Description} \\ \hline
        Closing duration $d_1$ & $t_2 - t_1$ \\ \hline
        Closed duration $d_2$ & $t_3 - t_2$ \\ \hline
        Reopening duration $d_3$& $t_4 - t_3$ \\ \hline
        Previous time    & \makecell[l]{Difference between this blinks $t_1$ to\\the previous blinks $t_1$.} \\ \hline
        Amplitude        & \makecell[l]{Relative amplitude, ratio of the difference\\of the maximum to the minimum ELA to\\the maximum ELA in the blink's window.} \\ \hline
        A/V ratio        & \makecell[l]{Ratio of $\mathrm{ELA}_{end} - \mathrm{ELA}_{min}$ to the\\maximum velocity in the reopening phase\\$m_2$.} \\ \hline
        Normal area $\mathrm{A_N}$    & \makecell[l]{Ratio of the shaded area $A$ to\\$(\mathrm{ELA}_{end}-\mathrm{ELA}_{min})\cdot 2d_3$. $A$ captures\\ the shape of the reopening phase.} \\ \hline
        PERCLOS \cite{wierwille_research_1994} & \makecell[l]{Ratio of the frames with $\mathrm{ELA}<\qty{20}{\degree}$ to\\the total number of frames between the\\end of this blink an the end of the\\previous blink.} \\ \hline
        PEROPENING & \makecell[l]{$\frac{d_3}{d_1+d_2+d_3}$} \\ \hline
    \end{tabular}
\end{table}

\subsubsection{Drowsiness detection}\label{sec:drowsiness-estimation}

Among others, Dreißig \textit{et al.}~\cite{dreisig_driver_2020} adopt the signal analysis framework proposed by Baccour \textit{et al.}~\cite{baccour_camera-based_2019} to classify driver states as \emph{awake} or \emph{drowsy}. 
Their method relies on a set of temporal and kinematic blink features, including the duration of the reopening phase, the amplitude–velocity ratio of the closing phase, the overall blink duration, and the blink frequency. 
Following this approach, we employ a \emph{$k$-Nearest Neighbors (kNN)} classifier with $k = 10$ to predict the drowsiness level based on features derived from the ELA signal.

Every $\qty{60}{\second}$, the drowsiness state is estimated using the blinking features computed within the current window. 
The complete set of features is listed in Table~\ref{tab:blinkfeatures}. 
To enhance feature robustness and comparability, we compute the mean and standard deviation of each feature across all detected blinks within the evaluation window. 
Each feature is then standardized to a normal distribution with $\mu = 0$ and $\sigma = 1$, followed by dimensionality reduction using \emph{Principal Component Analysis (PCA)} with five components.%

\subsection{Synthetic videos based on the ELA}
We created a controllable three-dimensional (3D) simulation environment using the open-source software \emph{Blender 3D}. 
The scene consists of a rigged human avatar \cite{clausell_yamoylse_nodate}, a single light source, and a virtual camera. 
Blender was chosen due to its extensive Python-based scripting API, which enables automated manipulation and rendering of scene parameters. 

The facial rig was configured to allow precise control of the eyelid angle according to a provided input signal, thereby enabling the generation of realistic blink dynamics. 
Additionally, the lighting conditions, as well as the position and orientation of the virtual camera, were programmatically adjustable to simulate different recording scenarios. 
For the experiments presented in this study, a single avatar model was used. 
An example of the rendered avatar is shown in Fig.~\ref{fig:teaser}. 

\subsubsection{Creation of new data}

Using the statistical distributions reported by Caffier \textit{et al.}~\cite{caffier_experimental_2003}, we synthesize numerical ELA signals representing individual blinks. 
Each blink is generated by constructing a continuous waveform composed of linear segments for the closing and reopening phases, and a spline-interpolated segment for the closed phase and its return to baseline. 
Blinks are concatenated using randomly sampled inter-blink intervals drawn from the empirical blink frequency distributions in~\cite{caffier_experimental_2003}. 
Because these distributions are provided separately for \emph{alert} and \emph{drowsy} subjects, the signal generation process can be conditioned on either state.

To emulate real-world variability, zero-mean random noise is added to the synthesized signal. 
Furthermore, the camera position is randomly determined using a heuristic normal distribution and linearly interpolated between set frames. The orientation is automatically adapted to always keep the avatar's head in the center of the frame.

\subsubsection{Creation of benchmark videos}\label{sec:synthbenchmarks}
A key advantage of the proposed Eyelid Angle (ELA) metric over alternative measures is its ability to be directly validated against a known ground truth. 
By explicitly controlling the parameters of the synthetic 3D scene, we generate a set of videos in which the ELA remains constant while the camera follows predefined trajectories around the avatar's head. 
This setup enables systematic evaluation of the ELA estimation accuracy under varying camera and face orientations.
For these validation experiments, no additional noise was introduced into the rendered data.


\subsection{Used datasets}

\subsubsection{Driver Monitoring Dataset}
We benchmarked our blink detector using the drowsiness section of the Driver Monitoring Dataset (DMD)~\cite{ortega_dmd_2020}, which provides frame-level annotations for blinks and closed-eye events such as microsleeps. 
Blinks are defined as the interval from eyelid closure onset to full reopening, consistent with our definition in Section~\ref{sec:blinkmetrics}. 
Some annotation inconsistencies were observed, including mislabeled or merged blink events (e.g. in Fig.~\ref{fig:blinkdetection}. 
The 16 analyzed videos contain $5441 \pm 183$ frames each at \qty{30}{\hertz}, totaling 1578 labeled blinks.

\subsubsection{UTA-RLDD}
The UTA-RLDD dataset~\cite{ghoddoosian_realistic_2019} contains about \qty{30}{\hour} of video categorized into \emph{alert}, \emph{low vigilant}, and \emph{drowsy} states based on the KSS scale~\cite{akerstedt_subjective_1990}. 
These classes are used to train our drowsiness detection network with features from preceding system stages. 
Several videos required manual correction of camera rotation for successful BlazeFace~\cite{bazarevsky_blazeface_2019} detection. 
Variable frame rates (VFR) in 169 of 182 videos were treated as constant at their mean FPS, and Participant~23 was excluded due to insufficient face size. 
Across 60 subjects, video quality varied in resolution, frame rate, and lighting, but remained consistent per subject.

\section{EXPERIMENTS AND RESULTS }

For our experiments, we first evaluate the ELA as a stable and reliable metric, then assess blink detection performance and finally drowsiness classification. 
We also analyze the influence of face rotation and frame rate using synthetic data.

\subsection{Comparison of ELA and EAR}
\begin{figure}[tb]
    \centering
    \vspace{2mm}
    \includegraphics[width=\linewidth]{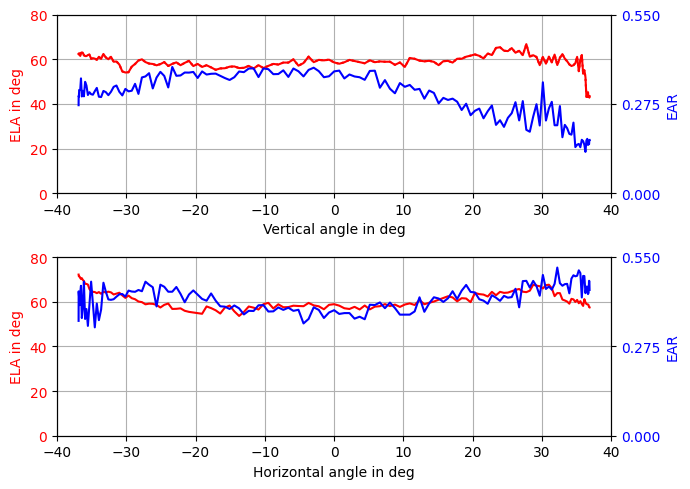}
    \caption{   
        Comparison of the EAR and raw ELA values across varying head orientations relative to the camera, measured on two synthetic videos generated as described in Section~\ref{sec:synthbenchmarks}. The vertical sweep ranges from below to above head level, and the horizontal sweep from right to left relative to the head. The eyelids were fixed at a constant angle of $\qty{60}{\degree}$.
. 
    }
    \label{fig:elaVSear}
\end{figure}

We compare the proposed Eyelid Angle (ELA) with the widely used EAR~\cite{soukupova_real-time_2016}. 
Both rely on 2D facial landmarks extracted from RGB images, unlike metrics such as EyeClosure~\cite{baccour_camera-based_2019}, which require specialized imaging setups. 
To study the effect of camera orientation, we render videos with fixed eyelid positions while sweeping the camera vertically and horizontally. 
An implementation following Soukupová and Čech~\cite{soukupova_real-time_2016} serves as the EAR baseline. 
As shown in Fig.~\ref{fig:elaVSear}, the raw ELA exhibits significantly lower variance across viewing angles, demonstrating greater robustness to orientation changes. 
This stability is partly due to MediaPipe’s improved landmark accuracy~\cite{king_dlib-ml_2009}, whereas the 2D nature of EAR makes it sensitive to projection effects under rotation.
Using synthetic videos with a constant ground-truth ELA of \qty{60}{\degree}, the measured ELA achieved mean absolute errors of \qty{2.8}{\degree} and \qty{3.3}{\degree} for vertical and horizontal sweeps, respectively.

\subsection{Error Across Ground Truth ELA and Face Orientation}
To evaluate the absolute error of the ELA under varying orientations and ground-truth values, we generated synthetic videos with set ELA values between \qty{0}{\degree} and \qty{70}{\degree} in \qty{10}{\degree} increments. 
Each video featured a scanning camera motion covering $\pm\qty{40}{\degree}$ vertical and horizontal rotations of the face. 
For each configuration, we computed the mean absolute error (MAE) and mean squared error (MSE) across all orientations.

As shown in Table~\ref{tab:ELAerror}, smaller ELA values yielded higher absolute errors. 
Extreme vertical rotations also increased deviations, although these are not explicitly listed in the table. 
Observed errors stem primarily from slight landmark placement inaccuracies in the eyelid regions and the heuristic computation of the 3D $z$-coordinates during landmark normalization.


\begin{table}[tb]
    \centering
    \caption{The error between the measured ELA and the set ground truth ELA changed over orientations and different set ELA, shown here. In general, the errors increased mostly over vertical rotations.}
    \label{tab:ELAerror}
    \begin{tabular}{|c||c|c|c|c|c|c|c|c|} 
        \hline
        \makecell[c]{set ELA\\in deg} &    0 &   10 &   20 &   30 &   40 &   50 &   60 &   70\\ \hline
        \makecell[c]{MAE\\in deg} &     18.3 & 14.6 & 10.8 & 13.1 & 11.6 &  7.6 &  4.1 &  6.1\\ \hline
        \makecell[c]{MSE\\in deg} &     442& 340& 194& 237& 171& 80 & 30 & 50\\ \hline
    \end{tabular}
\end{table}

\subsection{Blink Detection}\label{sec:detectionaccuracy}
The blink detector was evaluated using the drowsiness section of the Driver Monitoring Dataset (DMD)~\cite{ortega_dmd_2020}, which includes frame-level blink annotations. 
As participants were instructed to exhibit various drowsy behaviors such as yawning and microsleeping, this dataset presents a wide range of challenging cases. 

\begin{figure}[tb]
    \centering
    \vspace{2mm}
    \includegraphics[width=\linewidth]{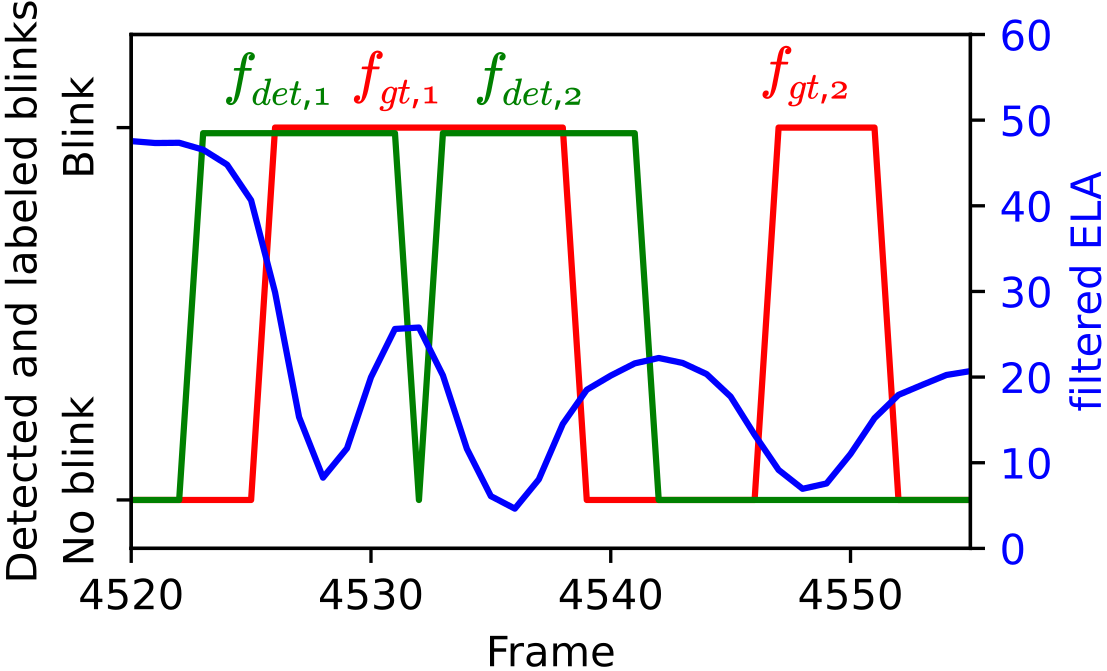}
    \caption{Exemplary detection results and ground-truth labels from video~7 of the DMD dataset. 
$f_{\mathrm{det},i}$ denotes the frames classified as part of a detected blink, and $f_{\mathrm{gt},i}$ the frames labeled as blinks. 
Shown are one false negative, where $f_{\mathrm{det},2}$ does not overlap with any labeled blink, and two true positives corresponding to a slightly misleading label spanning two actual blinks. 
In this example, the subject squints, yawns, and blinks multiple times.
}
    \label{fig:blinkdetection}
\end{figure}

Given the broad blink definition in DMD, we consider a detection by our system successful if a detected blink $f_\mathrm{det} \in B_\mathrm{det}$ overlaps with a labeled blink window $f_\mathrm{gt} \in B_\mathrm{gt}$. 
In contrast, detections outside labeled intervals are counted as false positives (FP), and labeled blinks without detections are counted as false negatives (FN). 
The detection accuracy (DA) is then defined as:


\begin{align*}
    \mathrm{TP} &= \{ f_{\mathrm{det}}\, | \,f_{\mathrm{det}} \cap f_{\mathrm{gt}} \neq \emptyset,\, \exists f_{\mathrm{gt}} \in B_\mathrm{gt},\, f_{\mathrm{det}}\in B_\mathrm{det}\},
    \\
    \mathrm{FN} &=  \{ f_{\mathrm{gt}}\, | \,f_{\mathrm{gt}} \cap f_{\mathrm{det}} = \emptyset,\, \forall f_{\mathrm{det}} \in B_\mathrm{det},\, f_{\mathrm{gt}}\in B_\mathrm{gt}\},
    \\
    \mathrm{FP} &=  \{ f_{\mathrm{det}}\, | \,f_{\mathrm{det}} \cap f_{\mathrm{gt}} = \emptyset,\, \forall f_{\mathrm{gt}} \in B_\mathrm{gt},\, f_{\mathrm{det}}\in B_\mathrm{det}\}, 
\end{align*}
as
\begin{align*}
    \mathrm{DA} &= \frac{\#\mathrm{TP}}{\#\mathrm{TP} + \#\mathrm{FN} + \#\mathrm{FP}} \cdot\qty{100}{\percent}
\end{align*}
where each labeled and detected blink is represented by its respective frame window. 
$TP$, $FP$, and $FN$ denote true positives, false positives, and false negatives, respectively, and $f$ represents the set of frames identified or labeled as part of a blink. 
An example of detection results and ground-truth labels is shown in Fig.~\ref{fig:blinkdetection}.


Our proposed system reached an average DA of \qty{89.4}{\percent} across all videos of the dataset.



\subsection{Drowsiness Detection}

\subsubsection{UTA-RLDD}
We inferred the ELA signal for each video in the UTA-RLDD dataset~\cite{ghoddoosian_realistic_2019} as a proof of concept to demonstrate the applicability of the proposed features. 
Our goal was not to achieve a new state of the art in drowsiness detection but to validate that ELA-based blink features can support such inference tasks.


Following the evaluation protocol defined by Ghoddoosian \textit{et al.}~\cite{ghoddoosian_realistic_2019}, subjects and their videos were divided into five folds, with four used for training and one for testing in rotation. 
Under this setup, our system achieved a video accuracy (VA) of \qty{52.5}{\percent}, below the \qty{65.2}{\percent} baseline reported in~\cite{ghoddoosian_realistic_2019} for the three-class classification problem. 
Given that our method relies solely on blink-related ELA features and a simple classifier, this result is expected. 
For the binary classification task (\emph{alert} vs. \emph{drowsy}), however, the VA improved to \qty{80.4}{\percent}. \cite{ghoddoosian_realistic_2019} did not give a VA score for their baseline value. \cite{dreisig_driver_2020} used a method similar to our proposal on a different dataset and report balanced accuracies of \qty{69.53}{\percent} for the multiclass and \qty{93.05}{\percent} for the binary classification problem.





\subsubsection{Synthetic data}
We generated four synthetic \qty{3}{\minute} ELA signals simulating blink behavior according to the metrics reported in~\cite{caffier_experimental_2003}, with added random noise. 
Two signals represented \emph{alert} and two \emph{drowsy} states, each rendered at \qtylist{10;30;50}{\hertz} to assess the influence of frame rate on blink and drowsiness detection. 
For each video, the ELA signal was inferred, blinks were detected, and blink features were extracted. 
The videos contained $48.5 \pm 0.7$ labeled blinks on average.

For each frame rate, the drowsiness classifier was trained on one \emph{alert} and one \emph{drowsy} video, and tested on the remaining pair at the same frame rate. 

The classification accuracy (AC1) and blink detection accuracy (DA) are summarized in Table~\ref{tab:syntheticdrowsiness}. 
Our method performs best around \qty{30}{\hertz}, while training a classifier across mixed frame rates degrades performance (AC2), as further analyzed in the following section.
Accuracy here is defined as the ratio of correctly classified clips of the previous \qty{60}{\second} for each detected blink of the videos.


\begin{table}[]
    \centering
    \caption{Influence of FPS on drowsiness detection accuracy when trained on videos with the same frame rate (AC1), trained across mixed frame rates (AC2), and the DA for synthetically generated videos.}
    \label{tab:syntheticdrowsiness}
    \begin{tabular}{|c||c|c|c|} \hline
        FPS & AC1 in \% & AC2 in \% & DA in \% \\ \hline
        10 & 77         & 69 & 51 \\ \hline
        30 & 98         & 69 & 95 \\ \hline
        50 & 92         & 46 & 95 \\ \hline
    \end{tabular}
\end{table}


\subsection{Stability of features over different frame rates}
There are several possible reasons why our $k$NN-based approach did not achieve the same accuracy as, for example, the implementation in~\cite{dreisig_driver_2020}. 
While~\cite{dreisig_driver_2020} employed a different dataset recorded at \qty{50}{\hertz} and extracted features from a metric with temporal characteristics similar to the ELA, our system was trained and evaluated on the UTA-RLDD dataset~\cite{ghoddoosian_realistic_2019}, which contains videos recorded at frame rates between \qtyrange{12}{30}{\hertz}. 
This discrepancy in sampling rate may therefore contribute to the reduced performance, as higher frame rates generally improve temporal signal fidelity.

To analyze the influence of frame rate, we generated synthetic videos based on the blink dynamics reported in~\cite{caffier_experimental_2003} and evaluated the effect on the features listed in Table~\ref{tab:blinkfeatures}. 
Most metrics asymptotically converged at higher frame rates; however, the detected \emph{closing duration} decreased by \qty{32}{\percent} between \qty{30}{\hertz} and \qty{50}{\hertz}. 
As this feature strongly discriminates between alert and drowsy states~\cite{caffier_experimental_2003,dreisig_driver_2020}, such sampling differences likely introduce systematic bias.

This frame rate dependence partially explains the discrepancies observed in Table~\ref{tab:syntheticdrowsiness} when training classifiers across videos recorded at different frame rates.

\section{CONCLUSIONS}

We introduced the ELA as a physically interpretable metric that quantifies the relative orientation between the upper and lower eyelids. 
Its applicability for generating synthetic data was demonstrated by recreating, augmenting, and extending existing datasets based on statistical blink distributions.

Future work will explore the generation of synthetic ELA signals using Generative Adversarial Networks (GANs), which could reduce the reliance on real-world training data and enhance the performance of visual drowsiness detection systems. 
The data generation framework presented here can support the training of various blink or drowsiness detectors, extending beyond landmark-based methods. 
Further research is needed to synthesize additional drowsiness-related behaviors such as yawning, establishing a foundation for comprehensive synthetic dataset creation.
With improving the generation of realistic synthetic drowsiness data, our proposed method also increases the amount of training data for transformer based drowsiness detectors like 

Current facial landmark detection systems already achieve impressive robustness, and continued improvements will further enhance the stability and accuracy of the ELA signal. 
While our implementation relied on a single uncalibrated camera and heuristic depth estimation (Eq.~\ref{eq:zheuristic}), alternative 3D landmark acquisition methods, such as multi-camera triangulation or depth-sensing cameras, could further reduce orientation-dependent errors (Fig.~\ref{fig:elaVSear}). 
Given that in-cabin driver monitoring systems must remain robust under large head rotations, multi-camera configurations represent a practical next step.
To facilitate reproducibility and future research, we will open-source the Blender~3D scene and data generation pipeline developed for this study, enabling straightforward adaptation to custom sensor or camera setups.

\bibliographystyle{ieeetr} 
\bibliography{DAAD_RISE_2025_ACFR_2}

\end{document}